\documentclass[journal]{IEEEtran}
\usepackage[pdftex]{graphicx}
\usepackage{multirow}
\usepackage{amsmath}
\usepackage{amssymb}
\usepackage{pifont}
\usepackage{caption}
\usepackage{makecell}
\usepackage{marvosym}

\newcommand{\xmark}{\ding{53}}%

\begin{document}
\title{CariMe: Unpaired Caricature Generation with Multiple Exaggerations}

\author{Zheng~Gu,
        Chuanqi~Dong,
        Jing~Huo,
        Wenbin~Li,
        and Yang~Gao
\thanks{Z. Gu, C. Dong, J. Huo, W. Li, Y. Gao are with the State Key Laboratory for Novel
Software Technology, Nanjing University, No. 163 Xianlin Avenue, Nanjing,
210023, Jiangsu, China. E-mail: guzheng@smail.nju.edu.cn, dongchuanqi@smail.nju.edu.cn, huojing@nju.edu.cn, liwenbin@nju.edu.cn, gaoy@nju.edu.cn.}}


\maketitle

\begin{abstract}
Caricature generation aims to translate real photos into caricatures with artistic styles and shape exaggerations while maintaining the identity of the subject. Different from the generic image-to-image translation, drawing a caricature automatically is a more challenging task due to the existence of various spacial deformations. Previous caricature generation methods are obsessed with predicting definite image warping from a given photo while ignoring the intrinsic representation and distribution for exaggerations in caricatures. This limits their ability on diverse exaggeration generation.
In this paper, we generalize the caricature generation problem from instance-level warping prediction to distribution-level deformation modeling. Based on this assumption, we present the first exploration for \textit{unpaired CARIcature generation with Multiple Exaggerations (CariMe)}. 
Technically, we propose a Multi-exaggeration Warper network to learn the distribution-level mapping from photo to facial exaggerations. This makes it possible to generate diverse and reasonable exaggerations from randomly sampled warp codes given one input photo.
To better represent the facial exaggeration and produce fine-grained warping, a deformation-field-based warping method is also proposed, which helps us to capture more detailed exaggerations than other point-based warping methods. Experiments and two perceptual studies prove the superiority of our method comparing with other state-of-the-art methods, showing the improvement of our work on caricature generation.
\end{abstract}


%
\IEEEpeerreviewmaketitle

\section{Introduction}

\IEEEPARstart{A} caricature is a pictorial representation of a person by exaggerating his/her most distinctive features in order to create an easily identifiable visual likeness~\cite{sadimon2010computer}.
Caricatures are often used in political satire as a ridiculous person or just for entertainment. 
Different from other non-photorealistic drawings like portraits~\cite{gatys2015texture}, sketches~\cite{yi2020unpaired} or cartoons~\cite{chen2018cartoongan}, caricatures are committed to reminding the viewer of the identity of the person through the use of pictorial hyperbole. 

Traditionally, artists draw caricatures by measuring the differences between the unique person and the average human face. 
While drawing caricatures, the artists are less bound by the constraints of reality, so that the facial features may be exaggerated beyond the possible. 
For example, an artist observes that someone's nose is bigger than average, so in caricature, the nose becomes likewise larger. 
With suitable exaggerations, the `likeness' of the caricatures may be even stronger than the real photos. However, the creation of a caricature usually takes great effort from a skillful artist, which is not convenient and friendly for others.
Therefore, it will be interesting and meaningful to automatically generate caricatures from a given real photo.

Automatically synthesizing a caricature from a real photo is not trivial in the field of computer vision.
Despite drawing the photo with a caricature texture style, we should also take spatial exaggerations into consideration~\cite{cao2018carigans,shi2019warpgan}. 
In fact, the exaggeration issue in faces is still an open problem in related researches ranging from detection~\cite{yaniv2019face}, recognition~\cite{shin2007combination} to generation~\cite{cao2018carigans}.
To address the exaggeration issue, some methods apply extra information such as user interaction~\cite{liang2002example} or 3D information~\cite{han2017deepsketch2face,wu2018alive} to guide the exaggerations.
Some methods implicitly achieve the exaggerations using deep neural networks in an image-to-image translation manner~\cite{han2018caricatureshop,zheng2019unpaired,li2020carigan}. 
There are also methods that learn point-based warping~\cite{cao2018carigans,shi2019warpgan} to translate real photos into caricatures. 

However, there are still two problems remain unsolved. 
First, most of these methods produce a definite exaggeration from an input photo in a one-to-one mapping manner. We summarize them as \textit{instance-level mapping} based methods. In the actual situation, however, the exaggerations of caricatures may differ among different professional artists~\cite{yaniv2019face}.
Therefore, the instance-level mapping which only searches a local suboptimal solution should be reconsidered.
Second, global affines or point-level sparse transformations mainly focus on changing overall facial contours, which makes the deformations more like global extrusions with less fine-grained local detail. This will make the deformations generated lack of reasonable local exaggerations. A latest method~\cite{gong2020autotoon} proposes to use the deformation field on caricature generation, but it requires paired warped photos created by professional artists, which is still laborious.

\begin{table*}[t]
\begin{center}
\caption{Comparison of related caricature generation methods}
\label{table_compare}
\begin{tabular}{|c|c|c|c|c|c|}
\hline
\textbf{Method} & \textbf{Warping Method} & \textbf{Supervision in Training} & \textbf{Supervision in Testing} & \textbf{Multiple Style} & \textbf{Multiple Exaggeration}\\
\hline
CariGANs~\cite{cao2018carigans} & point-based & 68 landmarks & 68 landmarks & ${\surd}$ & \xmark \\
\hline
WarpGAN~\cite{shi2019warpgan} & point-based & identity \& pre-trained classifier & none & ${\surd}$ & \xmark \\
\hline
AutoToon~\cite{gong2020autotoon} & deformation field & paired image & none & ${\surd}$ & \xmark \\
\hline
ours & deformation field & 17 landmarks & none & ${\surd}$ & ${\surd}$ \\
\hline
\end{tabular}
\end{center}
\end{table*}

In this paper, we present a novel multi-exaggeration caricature generation framework to tackle the problems above.
Different from previous caricature generation methods, 
we assume the exaggeration patterns can be described as a learnable high dimensional distribution which can be represented by a low-dimensional warp code. Based on this assumption, we model the deformations in caricatures by learning a \textit{distribution-level mapping} from the latent space to the exaggeration space. Given one input photo, random warp codes are sampled from a normal distribution to produce multiple exaggerations for this photo. 
Furthermore, to overcome the problem that affine transformation or point-based warping cannot produce meaningful fine-grained warping, our method is based on learning deformation field which can produce fine-grained image warping for caricature generation from unpaired data.
A deformation field is a position map specifying the sampling locations for each pixel in an image. Image transformations can be effectively conducted by any sampling method using a deformation field.

In addition, to achieve photo-specific deformation, a content code is also introduced to extract content information from input photos. During test, a Warper network takes a random warp code along with the content code as inputs to produce a meaningful exaggeration. By changing the warp codes, we can obtain multiple deformations while maintaining the identity of the given photo. 

The main contributions of this paper can be summarized as follows:  
\begin{itemize}
    \item We present a novel \textbf{CARI}cature generation framework with \textbf{M}ulti \textbf{E}xaggerations, CariMe, which can generate caricatures with both multiple exaggerations and multiple styles. This is the first work on multiple-exaggeration caricature generation.
    \item We propose a novel warping method based on deformation fields for unpaired caricature generation, which can effectively learn the spatial transforms distribution from real photos to caricatures. An auxiliary content code is also introduced to help our method to produce meaningful and photo-specific exaggerations.
    \item Experiments along with two perceptual studies show that CariMe can not only generate diverse exaggerations but also produce high-quality caricatures with better identity preservation and runtime performance, which shows the superiority of our method.
\end{itemize}

\begin{figure*}[t]
\centering
\includegraphics[width=1.0 \linewidth]{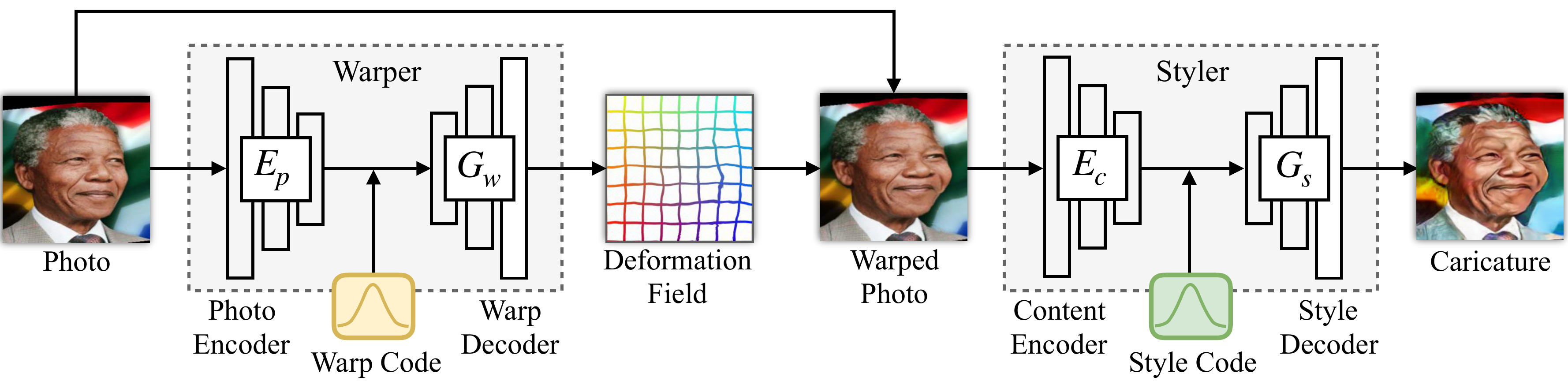}
\centering
\caption{\small The whole framework of our CariMe. The Warper first generates a deformation field for the given photo with a random sampled warp code to perform the exaggeration. After that, the Styler translates the warped photo into a caricature with the texture style controlled by a random style code.}
\label{fig_whole}
\end{figure*}

\section{Related Work}
In this section, we introduce three kind of related work in this paper: geometric warping generation, neural style transfer and automatic caricature generation.

\subsection{Geometric Warping Generation}
There has been a rich amount of research on generating geometric warping in visual and media tasks~\cite{lin2012patch,jin2016region}. These works can be divided into three categories: parameter-based, point-based, and field-based. Parameter methods~\cite{jaderberg2015spatial,lin2018st} apply a global transformation on features via a set of parameters. 
Point-based methods~\cite{cole2017synthesizing,wu2019transgaga} learns to predict critical control points(\emph{e.g.,} facial landmarks) to guide the geometric transformation between two image domains.
However, due to the limitation of the expressiveness of few parameters or sparse control points, these methods pay their most attention to global deformations and ignore detailed transforms in local areas.
Field-based methods directly generate the deformation field which is used to perform sampling to produce image transformation.
DeepWarp~\cite{ganin2016deepwarp} performs coarse-to-fine warping of individual pixels to gaze redirection in a small region like eyes. 
Deformable VAE~\cite{xing2018deformable} learns displacement for the coordinates of each pixel and performs geometric warping, such as stretching and rotation. 
These methods mainly focus on generating deformation to avoid visual distortion and preserve disparity consistency~\cite{fan2019stereoscopic}, rather than generating diversified deformation especially. In contrast to these methods, we aim to learn the rules of exaggerations to generate caricatures. Our model learns global deformations that can represent exaggerations with fine-grained detail in caricature, which is naturally more suitable for caricature generation task.

\subsection{Neural Style Transfer}
Style transfer is one kind of image synthesis problem which aims to render a content image with different styles~\cite{jing2019neural}. Gatys \textit{et al.}~\cite{gatys2015texture} make the first exploration in neural style transfer by extracting hierarchical features from a VGG network~\cite{simonyan2014very} and rendering styles through an optimization process.
WCT~\cite{li2017universal} and Adaptive instance normalization (AdaIN)~\cite{huang2017arbitrary} represents image style as second-order statistics of deep features and replace the optimization process with direct feature transformation.
Adaptive layer-instance normalization (AdaLIN)~\cite{kim2019ugatit} makes improvements over AdaIN by combining instance normalization and layer normalization together in the style stripping process.
Recently, with the great ability of generative adversarial networks (GANs)~\cite{goodfellow2014generative} to fit a data distribution, quite a lot of GAN-based methods are proposed~\cite{liu2019swapgan,zhou2019branchgan}. UNIT~\cite{liu2017unsupervised} and MUNIT~\cite{huang2018multimodal} assume a shared latent space across images. Upon this assumption, they can generate multiple styles from a common latent space. CycleGAN~\cite{zhu2017unpaired} achieves unpaired image translation with a cycle consistency loss. StarGAN~\cite{choi2018stargan} and StarGAN-v2~\cite{choi2020stargan} learn mappings among multiple image domains with one single generator.
However, directly learning the photo-to-caricature mapping in an image-to-image translation manner makes it hard for the model to capture the geometric transformation. In our method, we introduce a multi-exaggeration Warper module. This module can generate diverse and reasonable deformation fields that can be applied to warping photos into caricatures.

\subsection{Automatic Caricature Generation}

There are several automatic caricature generation works proposed before. CariGANs~\cite{cao2018carigans} first disentangles the generation into two sub-models. The first is a CycleGAN-based model (CariGeoGAN) which is trained on dense landmarks in the PCA subspace to perform a one-to-one geometry deformation from photo landmarks to caricature landmarks. The second is an MUNIT-based model (CariStyGAN), which transfers the texture to non-photorealistic caricature style. 

WarpGAN~\cite{shi2019warpgan} gather the warping and styling period in one single model. This method directly predicts a set of control points rather than facial landmarks. The control points warp a photo into a caricature without the supervision of landmarks. Besides, WarpGAN applies an identity-preservation adversarial loss to maintain the identity, this may bring one limitation that the discriminator needs to be pre-trained on an auxiliary dataset to obtain the ability to identify.

AutoToon~\cite{gong2020autotoon} uses deformation field to implement the exaggerations, which is very close to our work. But our framework is different from AutoToon in two aspects. First, AutoToon needs artist-warped photos to supervise the learning of warping field, which involves substantial skills and hard to obtain. Secondly, same with CariGANs and WarpGAN, AutoToon considers the generation problem as a one-to-one mapping, thus it also cannot generate diverse exaggerations. 

In this paper, we follow the assumption~\cite{cao2018carigans,gong2020autotoon} that disentangles caricature generation into two steps: texture style transfer and shape exaggeration. However, we also assume the exaggerations should be represented a distribution-level mapping from a latent code distribution to a deformation distribution.
This endows our model the ability to generate diverse exaggerations by sampling warp codes, and to produce fine-grained image warping with the identity preserved.

\begin{figure*}[t]
\centering
\includegraphics[scale=0.265]{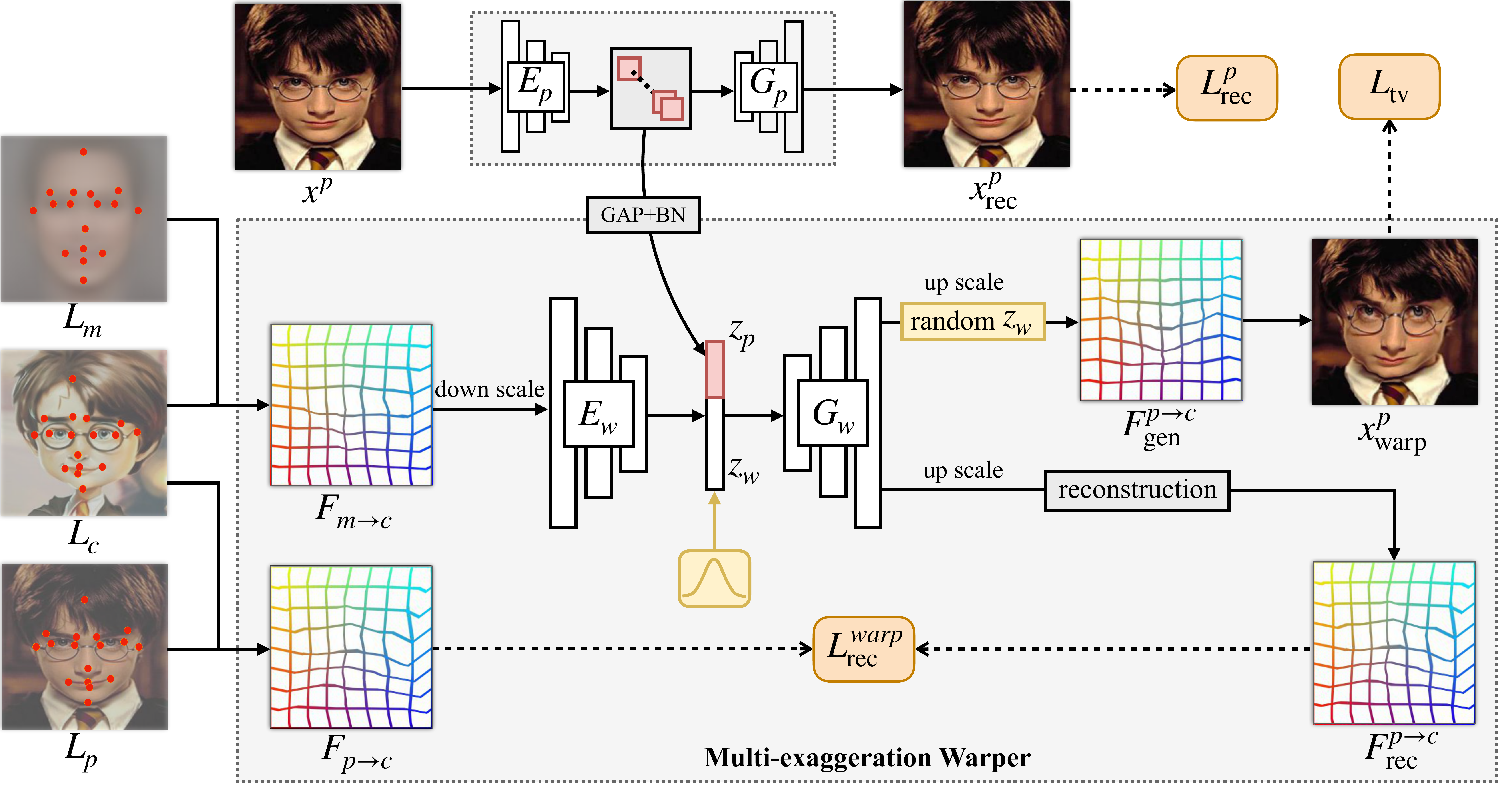}
\caption{The training phase of the Multi-exaggeration Warper. The Warper is encouraged to generate the ground truth deformation field from the warp code and content code, while producing smooth exaggeration from a random warp code.}
\label{fig_warper}
\end{figure*}

\begin{table}[t]
\begin{center}
\caption{Illustration for mathematical notations}
\label{table_notation}
\begin{tabular}{llll}
\hline\noalign{\smallskip}
Name& Meaning& Name& Meaning\\
\noalign{\smallskip}
\hline
\noalign{\smallskip}
$x^p$& real photo& $E_p$& photo encoder\\
$x^c$& real caricature& $G_p$& photo decoder\\
$L_p$& photo landmark& $E_w$& warp encoder\\
$L_c$& caricature landmark& $G_w$& warp decoder\\
$L_m$& mean landmark& $E_c$& content encoder\\
$z_s$& random style code& $E_s$& style encoder\\
$z_w$& random warp code& $G_s$& style decoder\\
$z_{p}$& photo content code& $D$& Discriminator\\
\hline
\end{tabular}
\end{center}
\end{table}

\section{The Proposed Method}

\subsection{Overall Framework}

Let $x^p$ be an image of real photo and $x^c$ be an image of real caricature. 
We assume the image can be disentangled into two latent representations $(C, S)$, where $C$ is the content representation capturing the spatial structure information and $S$ maintains the style information which is often reflected by texture and color\cite{yu2019multi}. 
The goal of CariMe to translate $x^p$ into a caricature with multiple geometric exaggerations along with various texture styles. 
To achieve multiple exaggerations, we propose a \textit{Multi-exaggeration Warper} module which takes a photo and a latent warp code as inputs to produce an exaggerated photo. 
For various texture, we propose a \textit{Styler} module which employs a latent style code to generate a caricature-like image while maintaining the spatial structure of this photo.
Given an input photo $x^p$, the geometric exaggeration is controlled by a warp code $z_w$, while the texture rendering controlled by a style code $z_s$. 
The whole framework of CariMe is shown in Fig.~\ref{fig_whole}. 
Important notations used in this paper are illustrated in Table \ref{table_notation}.

\begin{figure*}[t]
\centering
\includegraphics[scale=0.206]{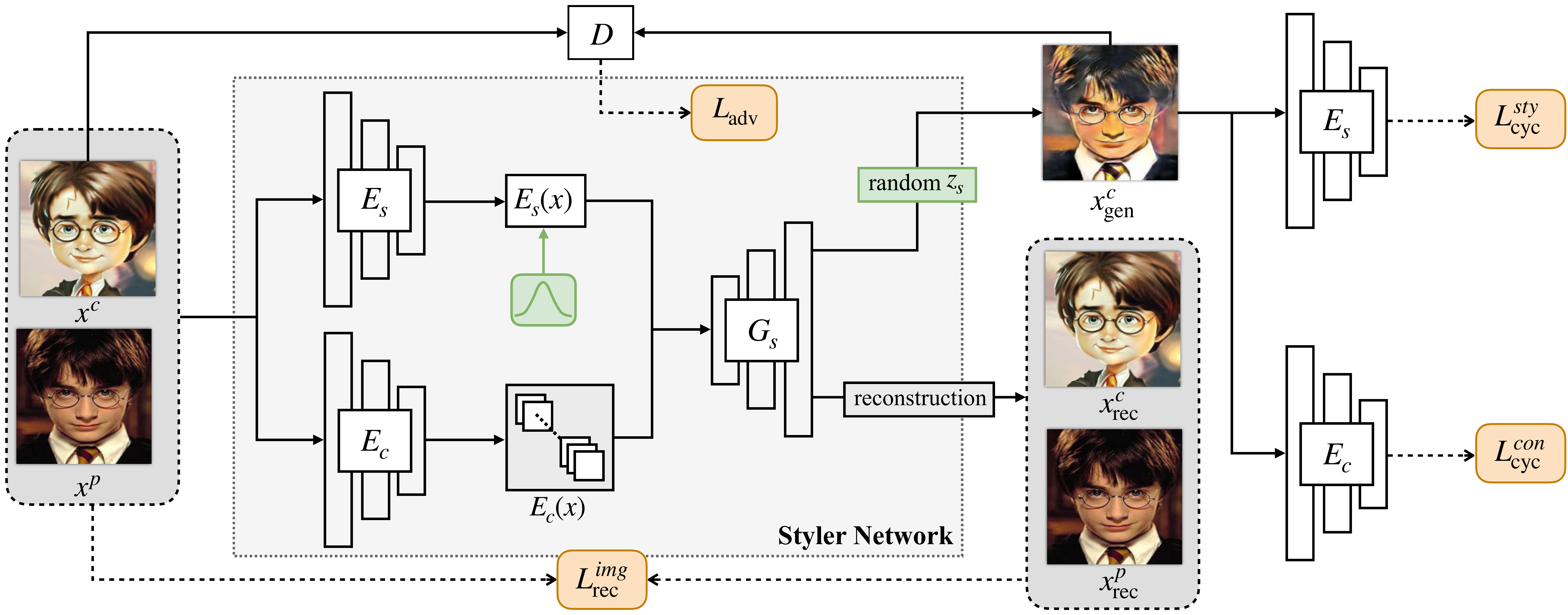}
\caption{The training phase of the Styler. The Styler should be able to reconstruct both the input photo and caricature, and meanwhile it tries to generate a fake caricature that the discriminator can not figure out from the real caricatures.}
\label{fig_styler}
\end{figure*}

\subsection{Multi-exaggeration Warper}
Given an input real photo, the purpose of Multi-exaggeration Warper is to generate diverse but meaningful deformations that can be used to perform the exaggerations on it. 
We start by using a latent warp code that follows the normal distribution to represent an exaggeration mode. 
Then with an extra content code encoded from the input photo, we obtain a deformation field that is conditioned on both the exaggeration and the input photo. The reason for combining both the warp code and content code is that, when drawing caricatures, we should consider not only the exaggeration but also the content of the photo (\textit{e.g.}, pose, expression).
In this case, we are trying to learn a mapping from a joint distribution of the geometric exaggeration and the image to the distribution of the deformation field. We define this as a \textit{distribution-level mapping} problem. 
The advantage of learning a distribution-level many-to-many mapping is that the distribution of the common exaggeration patterns can be better represented, making it possible to generate various deformations.

Fig.~\ref{fig_warper} shows the detailed architecture of the Warper module.
Let $L_p$ and $L_c$ denote the facial landmarks of the real photo and caricature, respectively.
We first calculate the mean landmarks $L_m$ of all the caricatures to represent a face with no exaggeration.
After that, a deformation field $F_{m \rightarrow c} \in \mathbb{R}^{W\times H\times 2}$ can be obtained for each caricature, which we respect to represent the specific exaggeration pattern in this caricature. 
In the training stage, a warp encoder $E_w$ encodes $F_{m \rightarrow c}$ into a normalized low-dimensional vector $z_w$, which is called a warp code. 
Meanwhile, we introduce an auxiliary photo-specific content code $z_p$ to capture content-aware information (\emph{i.e.,} pose, expression, identity) for the input photo $x^p$. This content code is obtained by synchronously training a photo encoder $E_p$ along with a photo decoder $G_p$.

In order to preserve the continuity the latent code space, the content code $z_p$ and warp code $z_w$ are first normalized to normal distribution, and then concated together to reconstruct the \textit{photo-specific deformation fields} $F_{p \rightarrow c}$ by a warp decoder $G_w$. 
In this way, a distribution-level bijection mapping is established between the exaggeration space and the latent space.
In the test stage, a random $z_w \sim \mathcal{N}(0, 1)$ is sampled with the same dimension with $z_w$ to obtain diverse exaggerations.
The deformation field is downscaled to $\mathbb{R}^{(W/2) \times (H/2) \times 2}$ before being fed into the warp encoder, and is upscaled to $\mathbb{R}^{W\times H\times 2}$ after the warp decoder. This operation makes the deformation field more smooth and noiseless.


There are three loss functions used in the Warper, \emph{i.e.,} a warp reconstruction loss $\mathcal{L}_{\mathrm{rec}}^{warp}$, a photo reconstruction loss $\mathcal{L}_{\mathrm{rec}}^{p}$, and a total variation loss $\mathcal{L}_{\mathrm{tv}}$.

\textbf{Warp Reconstruction Loss.}
Given a photo $x^p$ and a caricature $x^c$, $F_{p \rightarrow c}$ denotes the deformation flow from $x^p$ to $x^c$. 
We have $z_w=E_w(F_{p \rightarrow c})$ and $z_{p}=E_{p}(x^p)$. 
Since $z_w$ encodes the exaggeration pattern and $z_{p}$ contains the content information, the warp decoder $G_w$ should be able to reconstruct $F_{p \rightarrow c}$.
\begin{equation}
\mathcal{L}_{\mathrm{rec}}^{warp}=\mathbb{E}_{x^p \in \mathcal{X}^{p} ,  x^c \in \mathcal{X}^{c} }\left[\left\| G_{w}(z_w, z_{p}) - F_{p \rightarrow c} \right\|_{1}\right]\,,
\end{equation}
where $\| \cdot \|_1$ denote the $\ell_1$ norm of a matrix.

\textbf{Photo Reconstruction Loss.} To encourage $z_p$ to maintain the content and spatial information in $x^p$, the input photo $x^p$ should be able to be reconstructed from $z_p$ so that the content code $z_p$ is awareness on the content of the photo. Since it is difficult to directly reconstruct images from low dimensional vectors, we use the feature maps produced by the last convolutional layer in the photo encoder to reconstruct the input photo.
\begin{equation}
\mathcal{L}_{\mathrm{rec}}^{p}=\mathbb{E}_{x^p \sim \mathcal{X}^p}\left[\left\| G_{p}(E_{p}^{-1}(x^p)) -x^p\right\|_{1}\right]\,,
\end{equation}
where $E_{p}^{-1}$ denotes the activation of last layer in $E_p$.

\textbf{Total Variation Loss.} To encourage the warp decoder $G_w$ to produce smooth exaggerations, we apply a total variation regularization term to constrain the variation between adjacent pixels in the image.  Given an input photo $x^p$ and a random warp code $z_{w} \sim \mathcal{N}(0, 1)$, a warped photo $I$ can be obtained by adjusting the deformation field $G_w(z_w, E_{p}(x^p))$ for this input photo using bi-linear sampling. The regularization term is formulated as follows:
\begin{equation}
\begin{aligned} \mathcal{L}_{\mathrm{tv}}(I) &= \sum_{i=1}^{H-1} \sum_{j=1}^{W} \sum_{k=1}^{3}\left(I_{i+1, j, k}-I_{i, j, k}\right)^{2} \\ &+\sum_{i=1}^{H} \sum_{j=1}^{W-1} \sum_{k=1}^{3}\left(I_{i, j+1, k}-I_{i, j, k}\right)^{2}\,. \end{aligned}
\end{equation}

The full objective of the Warper is as below:
\begin{equation}
\mathcal{L}_W = \lambda_{\mathrm{rec}}^{warp} \mathcal{L}_{\mathrm{rec}}^{warp} 
+ \mathcal{L}_{\mathrm{rec}}^{p} 
+ \lambda_{\mathrm{tv}} \mathcal{L}_{\mathrm{tv}}\,,
\end{equation}
where $\lambda_{\mathrm{rec}}^{warp}$ and $\lambda_{\mathrm{tv}}$ are hyper-parameters.

\subsection{Styler}
The Styler aims to translate an input image to a caricature. Inspired by recent work on image-to-image translation~\cite{xiao2018elegant,gu2019mask}, we disentangle the image into a content representation $c$ and a style code $s$. Our Styler contains three modules: a content encoder $E_c$, a style encoder $E_s$ and a style decoder $G_s$. Given an input image $x\in\mathbb{R}^{W \times H \times C}$, $E_c$ encodes $x$ into a spatial feature map $E_c(x)$ which retains the spatial structure information, while $E_s$ encodes $x$ into a style vector $E_s(x)$ which represents the image style. For photo-to-caricature translation, a latent style code $z_s$ is sampled from a normal distribution, then $E_c(x^p)$ is processed by $G_s$ to get a caricature-like image with the style controlled by $z_s$. The overview of the Styler is shown in Figure \ref{fig_styler}.

We extend AdaLIN~\cite{kim2019ugatit} to implement our style decoder. Let $a$ denote the feature map produced by the content encoder $E_c(x)$, both instance normalization and layer normalization are performed to get two normalized features $a_\text{IN}$ and $a_\text{LN}$. We combine them together with a learnable weight $\rho$ and use several fully connected layers to predict the de-normalizing parameters $(\gamma, \beta)$ from the style code $z_s$.
\begin{equation}
AdaLIN(a) = \gamma \cdot (\rho \cdot a_{\text{IN}} + (1 - \rho) \cdot a_{\text{LN}}) + \beta\,.
\end{equation}

After several residual blocks with AdaLIN~\cite{kim2019ugatit}, the style decoder $G_s$ produces a caricature-like image from the processed feature map without exaggeration.

We use three loss functions to optimize the Styler, \emph{i.e.,} an adversarial loss $\mathcal{L}_{\mathrm{adv}}$, an image reconstruction loss $\mathcal{L}_{\mathrm{rec}}^{img}$, and a cycle consistency loss $\mathcal{L}_{\mathrm{cyc}}$.

\textbf{Adversarial Loss.} We employ adversarial loss to encourage the Styler to generate fake caricatures that the discriminator $D$ can not figure out from real ones. 
Here we adopt the Least Squares GAN's~\cite{mao2017least} objective.
\begin{equation}
\begin{aligned} \mathcal{L}_{\mathrm{adv}} &= \mathbb{E}_{x_{c} \in \mathcal{X}^{c} } [ (D(x_{c}))^2 ] \\ &+ \mathbb{E}_{x^{p} \in \mathcal{X}^{p}, z_s \sim p(s)}[(1-D(G_{s}(E_{c}(x^{p}), z_s)))^2] \,.
\end{aligned}
\end{equation}

\textbf{Image Reconstruction Loss.} For the module of style transfer, given an image from either photo domain or caricature domain, we reconstruct it after encoding and decoding.
\begin{equation}
\mathcal{L}_{\mathrm{rec}}^{img}=\mathbb{E}_{x \in \left( \mathcal{X}^{p} \cup \mathcal{X}^{c} \right) }\left[\left\|G_{s}\left(E_{c}\left(x\right), E_{s}\left(x\right)\right)-x\right\|_{1}\right]\,.
\end{equation}

\textbf{Cycle Consistency Loss.} For the generated image $G_s(E_c(x^p), z_s)$ with a random style code $z_s$, the style encoder $E_s$ should be able to encode the style code from the generated image, and the content encoder $E_c$ should be able to encode the content feature again.
\begin{equation}
\begin{aligned}
\mathcal{L}_{\mathrm{cyc}} &= \mathcal{L}_{\mathrm{cyc}}^{con} + \mathcal{L}_{\mathrm{cyc}}^{sty} \\ & = \mathbb{E}_{x^{p} \in \mathcal{X}^{p}, z_s \sim p(s)}\left[ {\left\| E_c(G_s(E_c(x^p), z_s)) - E_c(x_p) \right\|}_{1} \right] \\ & + \mathbb{E}_{x^{p} \in \mathcal{X}^{p}, z_s \sim p(s)}\left[ {\left\| E_s(G_s(E_c(x^p), z_s)) - z_s \right\|}_{1} \right] \,.
\end{aligned}
\end{equation}

The full objective of the Styler is as below:
\begin{equation}
\mathcal{L}_S = \mathcal{L}_{\mathrm{adv}} + \lambda_{\mathrm{rec}}^{img} \mathcal{L}_{\mathrm{rec}}^{img} + \lambda_{\mathrm{cyc}} \mathcal{L}_{\mathrm{cyc}}\,.
\end{equation}

\begin{figure*}[t]
\centering
\includegraphics[width=1.0 \linewidth]{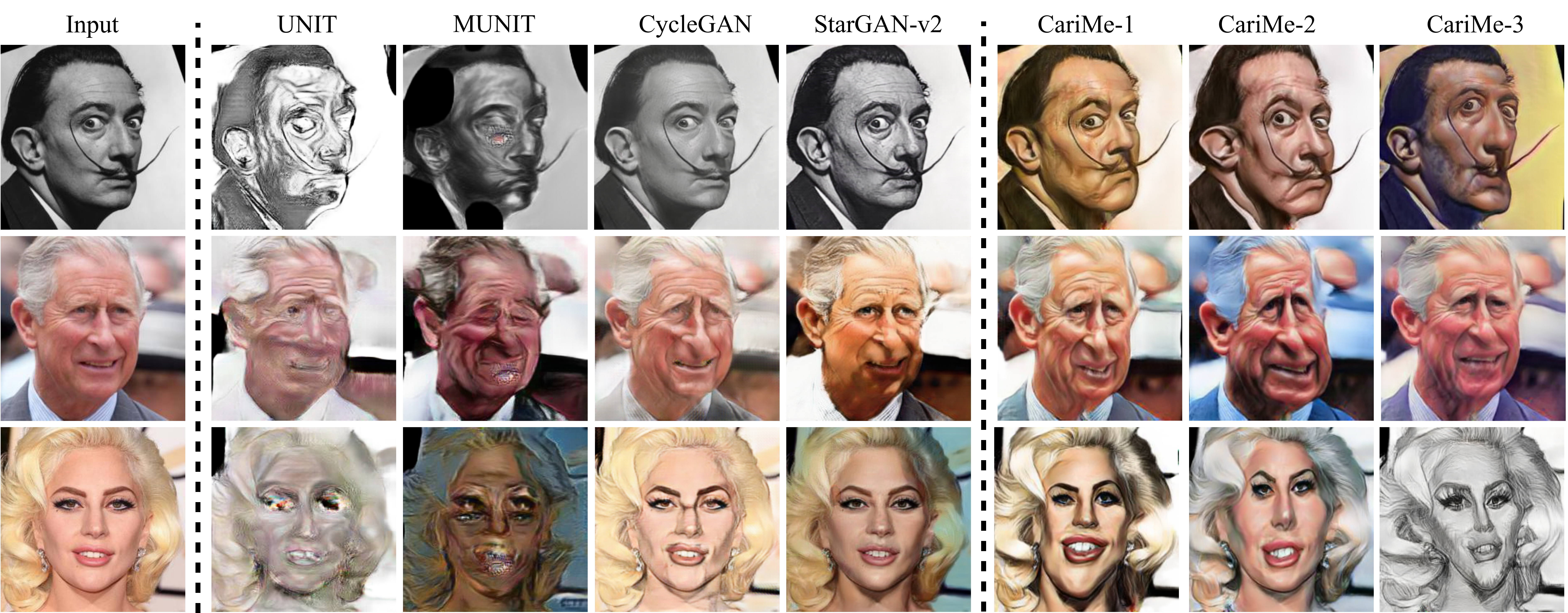}
\centering
\caption{\small Comparison of the proposed CariMe with general image-to-image translation methods. The last three columns show the results of our CariMe using different warp codes and style codes sampled from the normal distribution.}
\label{fig_sota_i2i}
\end{figure*}

\begin{figure*}[t]
\centering
\includegraphics[width=1.0 \linewidth]{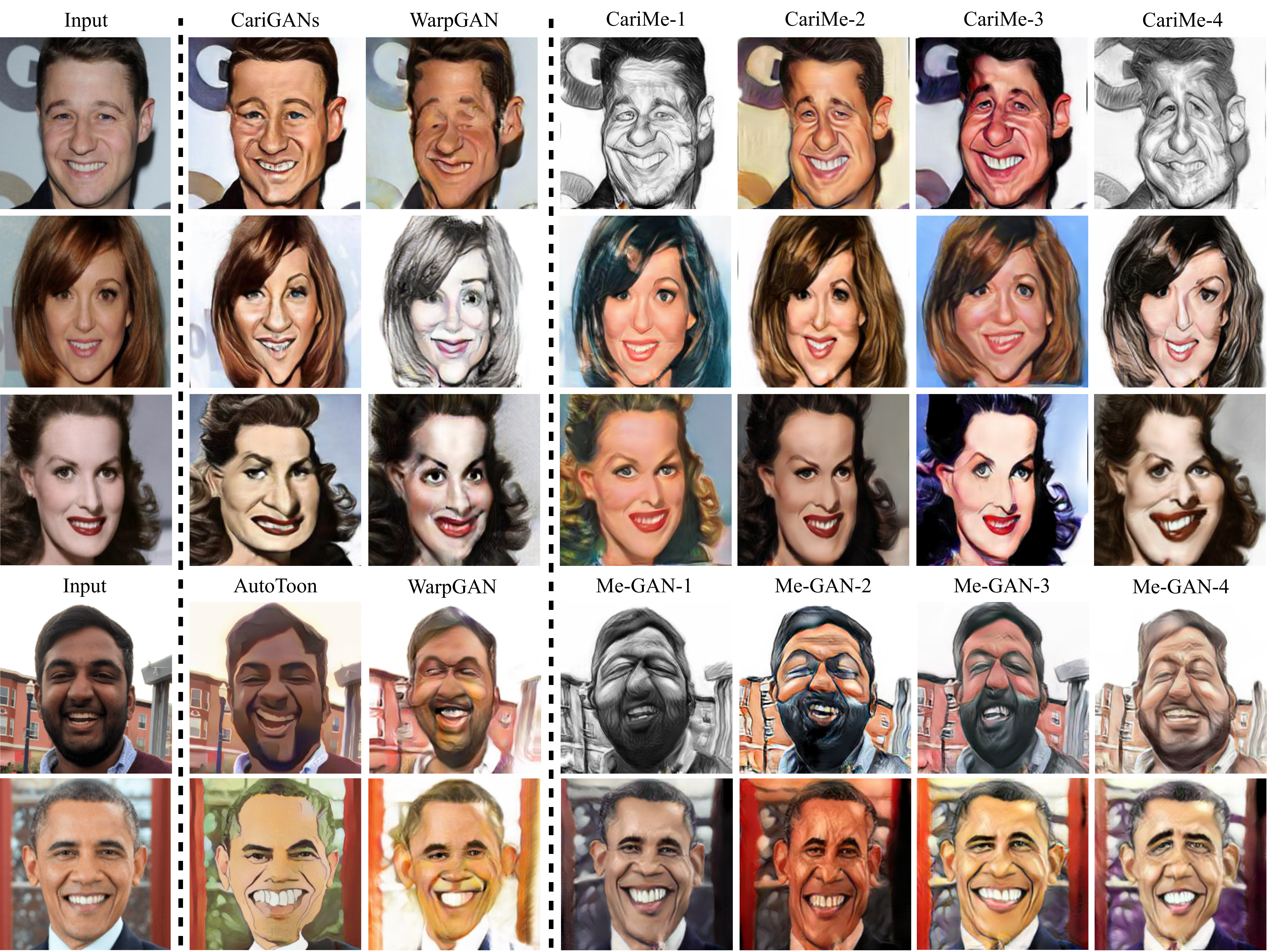}
\centering
\caption{\small Comparison of the proposed CariMe with other state-of-the-art caricature generation methods with both exaggeration and style rendering. The input and output images of CariGANs and AutoToon are selected from their papers. Our CariMe presents various output caricatures with different style and exaggeration while maintaining the identity well.}
\label{fig_sota_cari-sw}
\end{figure*}

\subsection{Training Details}
We train all of our networks using the Adam optimizer~\cite{kingma2014adam} with $\beta_1=0.5$ and $\beta_2=0.999$. For data augmentation, we horizontally flip the images, resize them to $288 \times 288$, and random crop them to $256 \times 256$ with a probability of $0.5$. 
The warper is trained using photo-caricature pairs from the same identity to learn more accurate warping patterns. 
The Styler is trained using photo-caricature pairs from randomly sampled identities (may be different) to get better style representation. 
The Warper is trained for $10,000$ iterations with a fixed learning rate of $0.0001$ and another $10,000$ iterations with the learning rate linearly decayed to 0. 
The Styler is trained with a fixed learning rate of $0.0001$ for $250,000$ iterations and another $250,000$ iterations with the learning rate linearly decayed to 0. 
As for the hyper-parameters, we set $\lambda_{\mathrm{rec}}^{img}=10$, $\lambda_{\mathrm{rec}}^{warp}=10$, $\lambda_{\mathrm{cyc}}=1$ and $\lambda_{\mathrm{tv}}=0.000005$.

\section{Experiments}

\subsection{Experimental Settings}
The experiments are performed on a large caricature dataset WebCaricature~\cite{HuoLSGY18}. This dataset contains $5,974$ photos and $6,042$ caricatures from $252$ identities. We use the 17 facial landmarks officially provided by the dataset to pre-process these images. To be specific, we first rotate each image by aligning the two eyes to the horizontal position according to four eye corners. Then, a bounding box can be obtained according to the face contour. 
Note that in many face recognition tasks, the images are usually resized without keeping the aspect ratio. But this is not suitable for generation tasks because the geometric distribution of human faces will be destroyed. Therefore, we also limit the bounding box to square to keep the aspect ratio of faces. 
After that, we enlarge the bounding box with a scale of $1.3$ in both width and height and resize it to $256\times256$. 
For data split, following WarpGAN~\cite{shi2019warpgan}, we randomly selected $126$ identities ($3,036$ photos and $3,005$ caricatures) for training and the remaining $126$ identities ($2,938$ photos and $3,037$ caricatures) for test.

\subsection{Comparison with Image-to-image Translation Methods}
First we compare our method with several state-of-the-art image-to-image translation methods, including UNIT~\cite{liu2017unsupervised}, MUNIT~\cite{huang2018multimodal}, CycleGAN~\cite{zhu2017unpaired} and StarGAN-v2~\cite{choi2020stargan}. All the methods are implemented using the officially released codes.

%
%
%

\begin{table}[!t]
\centering
\caption{Comparison on Fréchet Inception Distance}
\begin{tabular}{lc}
\hline 
\textbf{Method} & \textbf{FID Score} \\
\hline
UNIT~\cite{liu2017unsupervised} & $40.84$ \\
MUNIT~\cite{huang2018multimodal} & $43.65$ \\
CycleGAN~\cite{zhu2017unpaired} & $43.96$ \\
StarGAN-v2~\cite{choi2020stargan} & $50.28$ \\
WarpGAN~\cite{shi2019warpgan} & $50.35$ \\
\hline
CariMe(ours) & $\mathbf{33.56}$ \\
\hline
\end{tabular}
\label{table_fid}
\end{table}

Fig.~\ref{fig_sota_i2i} shows the results of our method compared with the above four general image-to-image translation methods. As can be seen, UNIT, MUNIT produce images with part of detailed semantic information lost. We believe this is because the large deformation gap between photos and caricatures makes the task harder for these methods. Although CycleGAN and StarGAN-v2 can generate clear outputs, the generated caricatures are somewhat identical to input photos with texture transfer but few deformations. The last three columns show the result of our proposed method. It can be seen that the output contains diverse facial exaggerations which are more like real caricatures. Table~\ref{table_fid} shows the comparison of Fréchet Inception Distance~\cite{heusel2017gans}, indicating that CariMe achieves a lower FID score than other methods.

\begin{figure*}[!t]
\centering
\includegraphics[width=0.95 \linewidth]{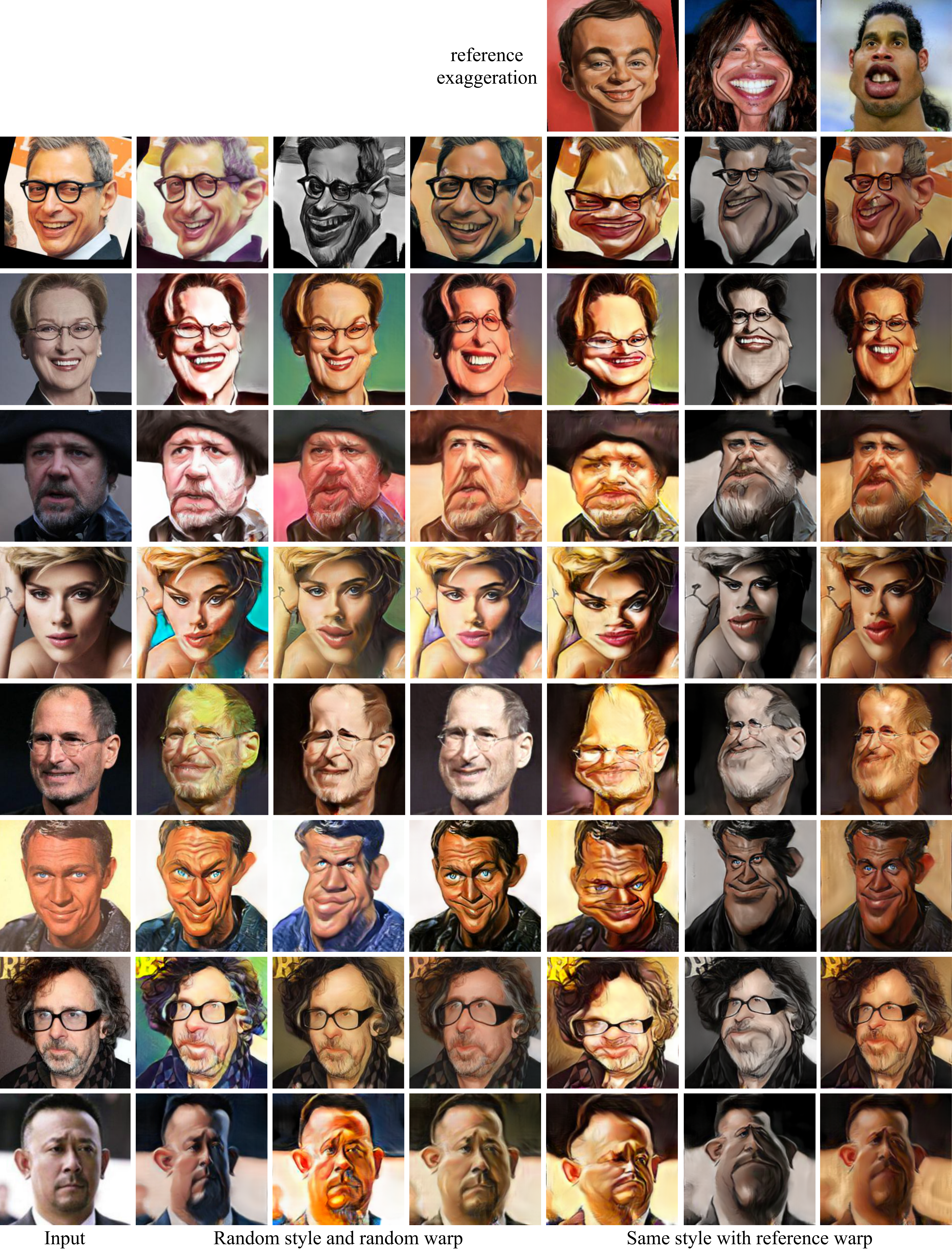}
\centering
\caption{Our method allow users to control on both style and exaggeration. Input photos are shown in the first column. The second to forth column show the results with random sampled style codes and warp codes. The last three columns show the results with the exaggeration guided by the reference exaggerations and the same style code.}
\label{fig_random_ref}
\end{figure*}

\subsection{Comparison with Caricature Generation Methods}
In this section, we make a comparison between our CariMe and three caricature generation methods CariGANs~\cite{cao2018carigans}, WarpGAN~\cite{shi2019warpgan} and AutoToon~\cite{gong2020autotoon}. We implement the warping module of CariGANs (\emph{i.e.,} CariGeoGAN) by ourselves. For WarpGAN, we use the officially provided code~\footnote{https://github.com/seasonSH/WarpGAN} and train the model using our training set. For AutoToon, we use the officially pretrained model~\footnote{https://github.com/adobe-research/AutoToon} as our baseline since we do not have paired data. We first take a brief introduction to these methods and make a discussion on the results. 

\textbf{CariGANs~\cite{cao2018carigans}} follows the structure of CycleGAN, but the inputs are vectors in the PCA space instead of images. After training, the model is used to predict the target landmark positions for the testing photos, then perform image warping on them from source facial landmarks to the target landmarks. We use the 17 landmarks provided by the WebCaricature dataset as supervision to train our CariGANs baseline.

\textbf{WarpGAN~\cite{shi2019warpgan}} uses one end-to-end model to generate both styles and warping at the same time. The style part is implemented using AdaIN~\cite{huang2017arbitrary}. The warping module of WarpGAN will predict a set of control points and their target positions for every input photo. The warping period is then performed using a sparse image warping method~\cite{cole2017synthesizing}. Moreover, the discriminator is pre-trained on an auxiliary dataset.

\textbf{AutoToon~\cite{gong2020autotoon}} learns a direct mapping from an input photo to a warping flow from paired data. The deformation field is required to make the warped photo to be close to the ground-truth warped photo that annotated by human beings. The stylization period of AutoToon is implemented using CartoonGAN~\cite{chen2018cartoongan}.

\begin{figure*}[t]
\centering
\includegraphics[width=1.0 \linewidth]{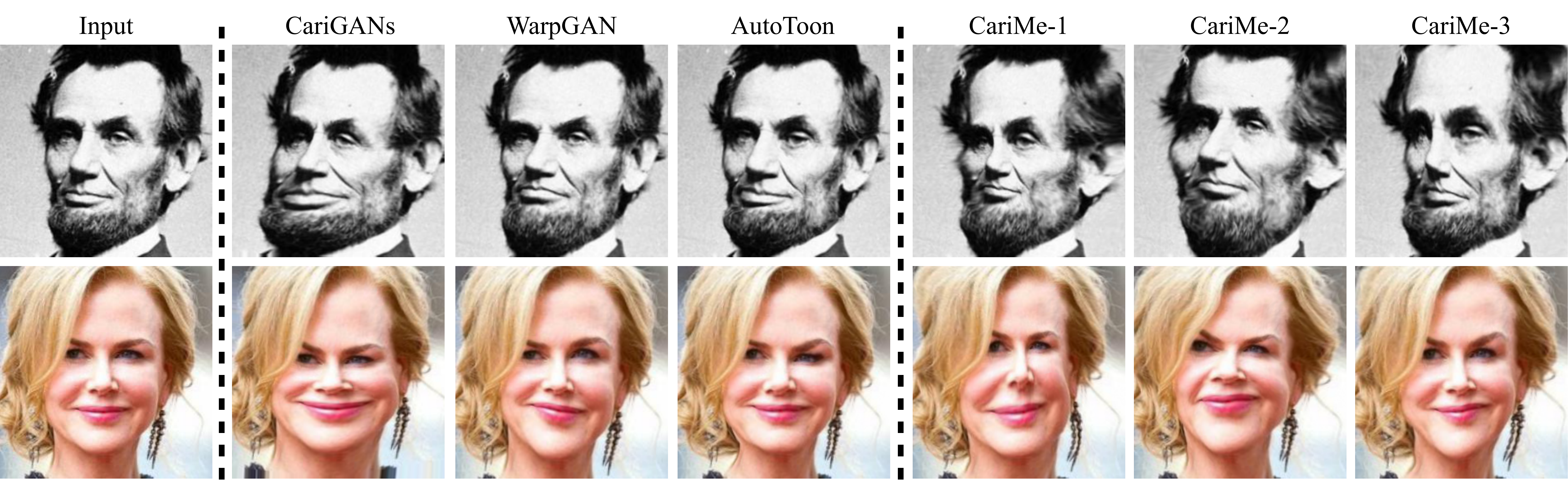}
\centering
\caption{\small Comparison of the proposed CariMe with other state-of-the-art caricature generation methods with only photo warping. Our CariMe can generated obvious and diverse exaggerations in the last three columns.}
\label{fig_sota_cari-w}
\end{figure*}

Fig.~\ref{fig_sota_cari-sw} shows our comparison with these methods. The input photos in the first three rows are from CariGANs paper and the last two rows are from AutoToon paper. The results for CariGANs and AutoToon are from their papers. As can be seen, the deformation of CariGANs~\cite{cao2018carigans} is more like some squeezes on the facial features. For example, the caricature generated by CariGANs in the third line is squeezed down, making the caricature not like the same person. WarpGAN~\cite{shi2019warpgan} may produce unclear artifacts in some local areas. As for AutoToon~\cite{gong2020autotoon}, the caricatures are less detailed and also present incoherent lines around the facial contour. The last four columns are caricatures given by our method. Each output is generated from a randomly sampled warp code and a randomly sampled style code. More results in our test set are given in Fig.~\ref{fig_random_ref}. We also make comparison with only image warping in Fig.~\ref{fig_sota_cari-w}. It can be seen that the outputs generated by CariMe are obviously in different exaggerations and the identity can also be kept. These results show the ability of our method to produce multiple and fine-grained exaggerations with fewer supervision than other methods.

\begin{figure*}[!t]
\centering
\includegraphics[width=0.99 \linewidth]{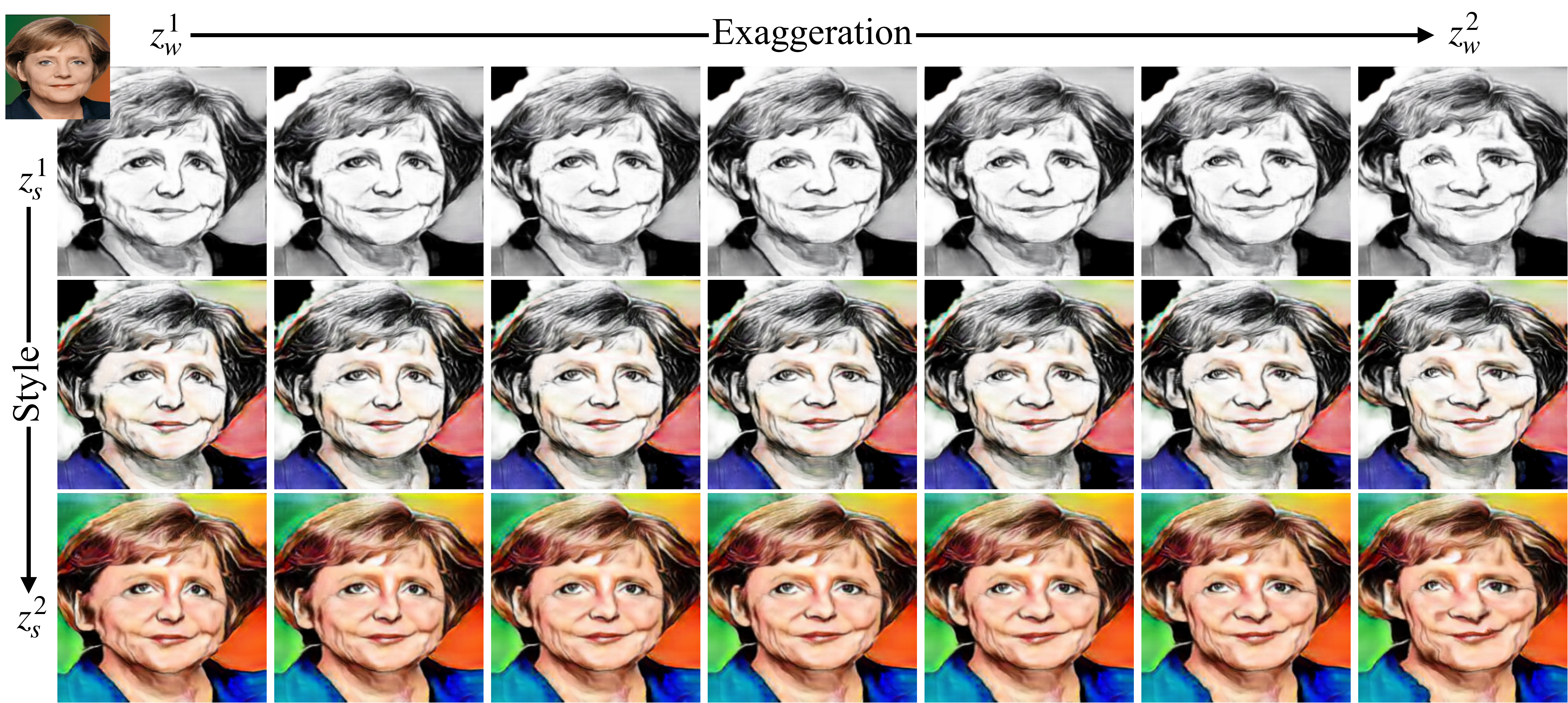}
\centering
\caption{Interpolation in both styles and deformations. The upper left corner is the input photo. The images in the same row share the same style, while the images in the same column have the same exaggeration.}
\label{fig_interpolation_sw}
\end{figure*}

\subsection{Controlling Both Style and Deformation}
Fig.~\ref{fig_interpolation_sw} exhibits the interpolation results with two randomly sampled pairs of style codes $(z_s^1, z_s^2)$ and warp codes $(z_w^1, z_w^2)$. Then we linearly interpolate style codes and warp codes inside to generate different outputs. When observing from top to bottom, the images are gradually colored (\emph{i.e.,} stylized). While observing from left to right, the exaggeration changes smoothly from one exaggeration pattern to another (\emph{e.g.,} a small nose becomes a big one). This shows the smoothness of the learned code space.

\begin{figure*}[!t]
\centering
\includegraphics[width=1.0 \linewidth]{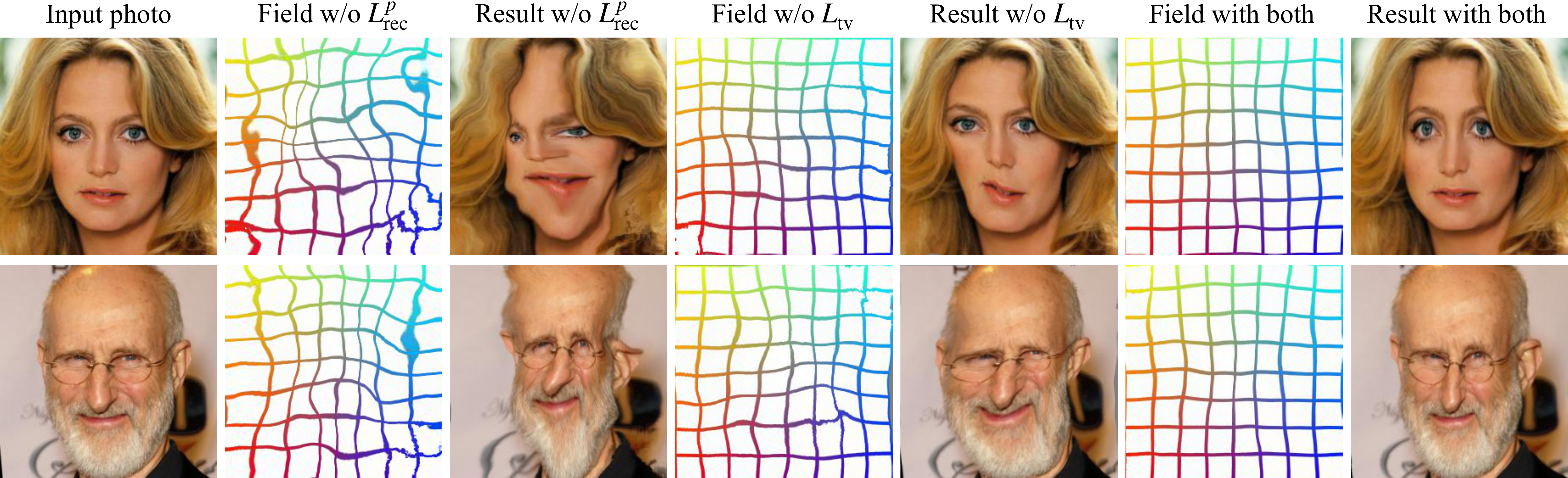}
\centering
\caption{The warped photos and deformation fields after removing different loss terms in the Warper. The Warper cannot produce meaningful exaggerations without neither content information nor total variation loss.}
\label{fig_ablation}
\end{figure*}

\begin{figure*}[!t]
\centering
\includegraphics[width=1.0 \linewidth]{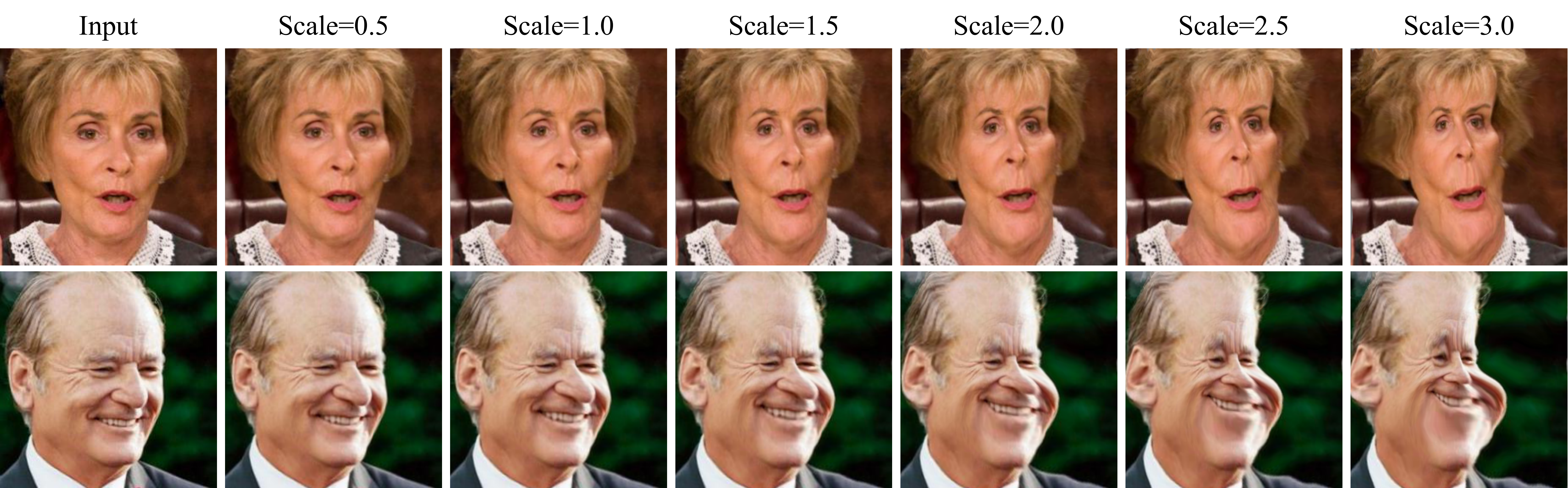}
\centering
\caption{\small Results with different scales on the deformation flow. In the first row, the exaggeration tends to present a longer face, smaller noise and smaller eyes with the scale increases. In the second row, it turns to present a larger chin, more extreme laugh style since the input photo is a smile one.}
\label{fig_interpolation_scale}
\end{figure*}

\subsection{Ablation Study}
For the newly designed Multi-exaggeration Warper in the proposed CariMe, an ablation study is performed to test the influence of different components. We remove several loss functions to demonstrate effectiveness. The results of removing $\mathcal{L}_{\mathrm{rec}}^{p}$ and $\mathcal{L}_{\mathrm{tv}}$ are reported in Fig.~\ref{fig_ablation}. Without the photo reconstruction loss $\mathcal{L}_{\mathrm{rec}}^{p}$, the deformation field is less bounded and is too distorted, and even cannot be used to produce meaningful exaggerations. Without the total variation loss $\mathcal{L}_{\mathrm{tv}}$, there appears some cracks appear in local areas. With both of these losses, our model can produce smooth and convincing exaggerations.

\subsection{Controlling Exaggeration Scale}
Same as previous methods like WarpGAN~\cite{shi2019warpgan} and CariGANs~\cite{cao2018carigans}, our method also allows customization on the exaggeration extent. This can be implemented by alternatively generating the residual deformation flow for each pixel in the sampling map. The deformation flow is multiplied by a scaling coefficient to control the exaggeration scale. Fig.~\ref{fig_interpolation_scale} shows the results of applying different scales on the input photo. With the scale increasing, the exaggeration becomes bigger while still being reasonable.

\section{Analysis}

\begin{table}[!t]
\centering
\caption{Comparison on Identity Preservation.}
\begin{tabular}{lcccc}
\hline
\textbf{Probe} & \textbf{Method} & \textbf{Warp Degree} & \textbf{Rank-1} \\
\hline
Real Photo & - & - & $94.80$ \\
Real Caricature & - & - & $60.87$ \\
\hline
\multirow{4}*{\makecell{Warped Photo\\(Small)}} & WarpGAN~\cite{shi2019warpgan} & $3.12$ & $93.76$ \\
& CariGANs~\cite{cao2018carigans} & $3.14$ & $\mathbf{94.18}$ \\
& AutoToon~\cite{gong2020autotoon} & $3.06$ & $90.96$ \\ 
& CariMe(ours) & $3.19$ & $\underline{94.06}$ \\ 
\hline
\multirow{4}*{\makecell{Warped Photo\\(Middle)}} & WarpGAN~\cite{shi2019warpgan} & $10.06$ & $82.40$ \\
& CariGANs~\cite{cao2018carigans} & $11.01$ & $\underline{84.69}$ \\
& AutoToon~\cite{gong2020autotoon} & $10.22$ & $60.97$ \\ 
& CariMe(ours) & $10.61$ & $\mathbf{87.60}$ \\ 
\hline
\multirow{4}*{\makecell{Warped Photo\\(Large)}} & WarpGAN~\cite{shi2019warpgan} & $30.01$ & $35.46$ \\
& CariGANs~\cite{cao2018carigans} & $32.33$ & $\underline{41.42}$ \\
& AutoToon~\cite{gong2020autotoon} & $25.54$ & $12.79$ \\ 
& CariMe(ours) & $31.73$ & $\mathbf{57.81}$ \\ 
\hline
\end{tabular}
\label{table_identification}
\end{table}

\subsection{Identity Preservation}
One concern about caricature generation methods is whether the identity can be preserved. In this section, we make a discussion on the identity preservation issue. 

We first propose two assumptions in this experiment. 
First, a good caricature should contain both large exaggeration and high accuracy(\emph{e.g.,} when the shape transformation is small, the accuracy should be high, but it may not be a good caricature). Identity preservation should be measured while keeping the degree of exaggeration be the same for different methods. 
Second, since the caricature styles may interfere with the evaluation of deformation quality, we only generate warped photos without caricature style transfer for every testing photo. 

Therefore, we should generate warped photos with the same degree of exaggeration by controlling the scale factor of different caricature generation methods. To achieve this goal, for each input photo $P \in \mathbb{R}^{W\times H\times 3}$, we first generate the corresponding deformation flow
 $F \in \mathbb{R}^{W\times H\times 2}$, then calculate the exaggeration degree for $P$ as follows:
\begin{equation}
Degree(P) = \frac{\sum_{i=1}^{W} \sum_{j=1}^{H} ||F_{i,j}||_2 }{H * W}\,.
\end{equation}

\begin{figure*}[!t]
\centering
\includegraphics[width=1.0 \linewidth]{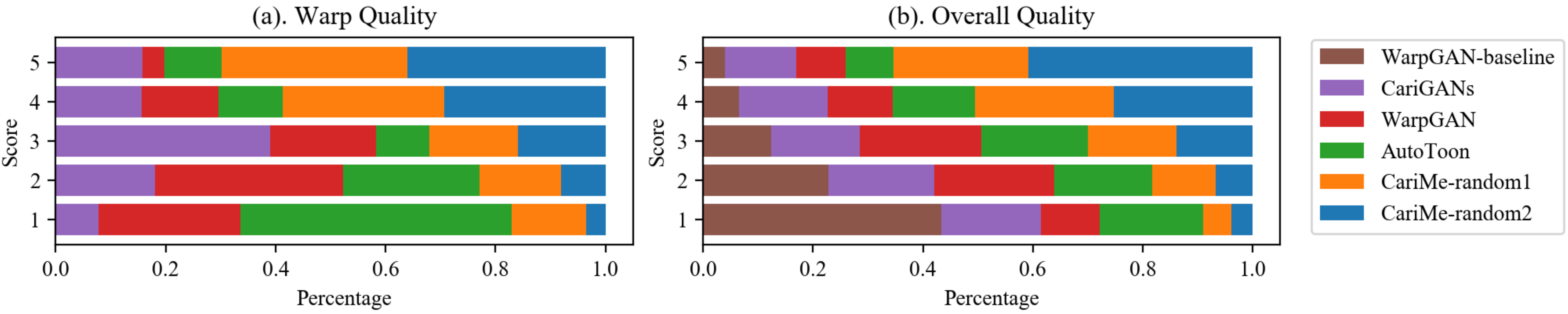}
\centering
\caption{Detailed result of perceptual study. (a) Percentage of each method that has been annotated by all volunteers in each score in the first warp quality study. (b) Percentage of each method that has been annotated by all volunteers in each score in the second overall quality study. Higher percentage in high scores reflects better user acceptance.}
\label{fig_user_study}
\end{figure*}

The warp degree for every method is calculated by averaging the degrees for all testing photos($2,938$ photos in total). This metric provides a unified standard for measuring the degree of exaggeration between different methods. 

After we obtain warped photos in the same warp degree, we use SE-ResNet~\cite{hu2018squeeze} pre-trained on VGG-Face2~\cite{cao2018vggface2} as a feature extractor and calculate the rank-1 accuracies. For each identity, we keep one photo in the gallery, while all the remaining photos, all real (hand-drawn) caricatures, and all generated warped photos are used as probes, respectively. 
The result for identity preservation is given in Table~\ref{table_identification}. We conducted three sets of experiments for different degrees of exaggerations: warped photo with small exaggeration(degree$\approx$3), middle exaggeration(degree$\approx$10), and large exaggeration(degree$\approx$30). The result shows CariMe can better preserve the identity than other methods, especially in large exaggerations. We also find that the scale factors required to uniform the degree are different among different methods. For example, for small exaggeration, CariMe needs a scale of $0.3$ to get warp degree $3.19$, while AutoToon needs a scale of $1.2$ to get warp degree $3.06$, which also proves our first assumption above.

\begin{table}[!t]
\centering
\caption{Average Scores for Perceptual Study.}
\begin{tabular}{lcc}
\hline 
\textbf{Method} & \textbf{Warp Quality} & \textbf{Overall Quality} \\
\hline
WarpGAN-baseline~\cite{shi2019warpgan}& - & $2.13$ \\
\hline
WarpGAN~\cite{shi2019warpgan}& $2.38$ & $2.81$ \\
CariGANs~\cite{cao2018carigans}& $3.04$ & $2.84$ \\
AutoToon~\cite{gong2020autotoon}& $2.15$ & $2.78$ \\
\hline 
CariMe-random1&$\underline{3.36}$ & $\underline{3.45}$ \\
CariMe-random2&$\mathbf{3.74}$ & $\mathbf{3.78}$ \\
\hline 
\end{tabular}
\label{table_user_study}
\end{table}

\subsection{Perceptual Study}

To further evaluate our method, we conduct two perceptual studies in terms of exaggeration and overall quality. $20$ volunteers are invited to our perceptual studies. The volunteers are first given 20 real caricatures to get familiar with real exaggerations, then asked to compare caricatures generated by our method with other caricature generation methods.

The first study is a warping quality test. In this study, we randomly choose $30$ photos with different identities in the test set. For each photo, we apply the warping module of different methods(\emph{i.e.,} CariGANs~\cite{cao2018carigans}, WarpGAN~\cite{shi2019warpgan}, AutoToon~\cite{gong2020autotoon}) to generate warped photos without stylization to avoid the influence of texture. Since our method is the only approach for multiple exaggerations, we use two random warp codes for every photo(\emph{i.e.,} CariMe-random1, CariMe-random2). However, the volunteers are told that the outputs are generated from five different methods. Then we ask the volunteers to score how much the exaggerations look like real caricature-style exaggerations. The scores range from 1 to 5 where a higher score means better warping. The results are reported in Table~\ref{table_user_study}, where our method achieves $3.36$ and $3.74$ for warp quality independently. We also present the percentage of each method in our perceptual study in Fig.~\ref{fig_user_study}, which shows a more detailed comparison.

The second study is an overall quality test. Similarly, we randomly choose another $30$ photos with different identities in the test set. 
We first produce warped photos using different caricature generation methods for each photo (same as the first study). Then we apply a stylization network(the Styler network in this paper) on these photos. We believe it will contribute to a more fair comparison by using the same stylization method. We also use the whole WarpGAN model as our baseline to evaluate the effectiveness of the Styler(WarpGAN-baseline). The scoring period is the same as the first study. The result is given in Table~\ref{table_user_study}, where our method outperforms on other methods. Moreover, we can see that for the same warped photo, applying the proposed Styler network can achieve better user acceptance($2.81$) than our baseline($2.15$).

\begin{table}[!t]
\centering
\caption{Running Time for Geometric Warping.}
\begin{tabular}{lcc}
\hline 
\textbf{Method} & \textbf{Time per Image} & \textbf{Time Faster} \\
\hline
CariGANs~\cite{cao2018carigans} & $0.006$s & $1.9\times$ \\
AutoToon~\cite{gong2020autotoon} & $0.013$s & $4.2\times$ \\  
WarpGAN~\cite{shi2019warpgan} & $0.024$s & $7.6\times$ \\ 
\hline 
CariMe(ours) & $\mathbf{0.003}$s & $-$ \\ 
\hline
\end{tabular}
\label{table_time}
\end{table}

\subsection{Performance}
Our core algorithm is developed in PyTorch 1.4~\cite{paszke2019pytorch}. All of our experiments are conduct on an NVIDIA V100 GPU. To evaluate the time cost for each method, we calculate the total time to generate warped photos for all testing photos (\emph{i.e.,} $2938$ photos) and get the average reasoning time. The runtime comparison is given in Table~\ref{table_time}. As can be seen, CariMe earns the first order in runtime performance, taking $0.003$ second to process one image. This is because the generated deformation field can be directly used to warp the photo by a simple sampling without other interpolate calculations like point-based methods. We also find that CariGANs take less time for geometric warping than AutoToon. This may be due to the use of fewer landmarks for CariGeoGAN in our experiment ($68$ to $17$). The upscaling of deformation fields in AutoToon from $32 \times 32$ to $256 \times 256$ also increases complexity.


\begin{figure}[!t]
\centering
\includegraphics[width=1.0 \linewidth]{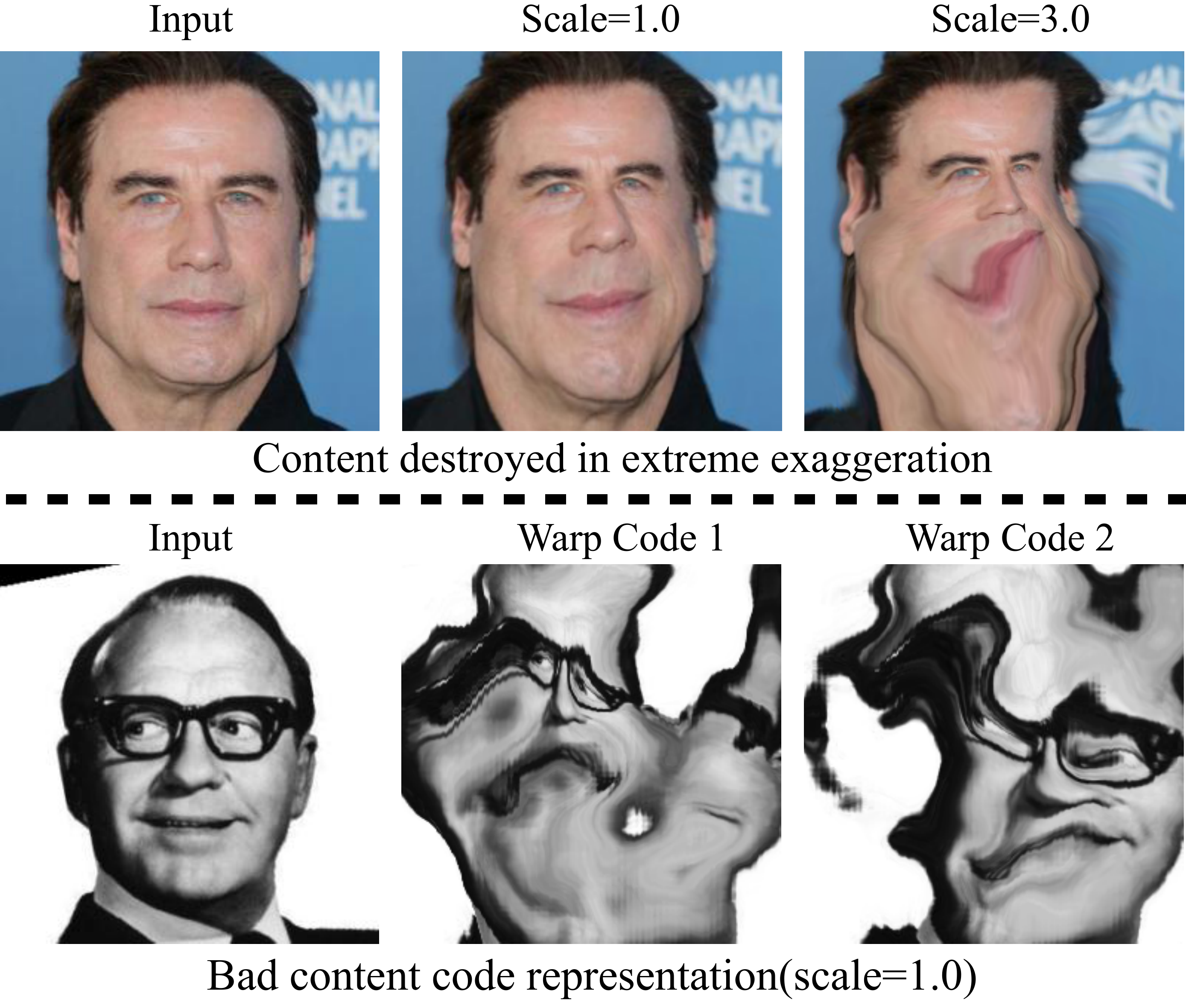}
\centering
\caption{\small Some failure cases. In the first line, the face is destroyed in some extreme exaggeration. In the second line, the output exaggeration is in chaos, which is caused by a badly learned content code.}
\label{fig_fail}
\end{figure}

\section{Conclusion And Future Work}
In this paper, we propose a novel framework for multi-exaggeration unpaired caricature generation. We propose a deformation field based warping method which considers both exaggeration pattern and image content to produce diverse caricature exaggerations. Our approach supports flexible controls to change shape exaggerations, warping degrees, and appearance styles. Experiments and perceptual studies demonstrate that the proposed method generates caricatures that are superior to other state-of-the-art caricature generation methods. 
Although our method can achieve compelling results in many cases, there still exist problems that need to be tackled in caricature generation. Fig.~\ref{fig_fail} shows several typical failure cases. The first line shows a failure example in large exaggerations where some content is destroyed, which also occurs in other caricature generation methods. The second line shows a failure example where the deformation field is in chaos no matter what warp code is sampled. We analyze this is caused by a badly learned content code for some isolated outliers. A possible way to tackle this problem is to enhance the representation ability of the photo encoder and feed it with more data. We will explore these interesting but unsolved problems in the future.

%


%

%
%
%
%
%

\ifCLASSOPTIONcaptionsoff
  \newpage
\fi



\bibliographystyle{IEEEtran}
\bibliography{IEEEabrv,egbib.bib}
\end{document}